\documentclass[letterpaper]{article} 
\usepackage{aaai2026}
\usepackage{times}  
\usepackage{helvet}  
\usepackage{courier}  
\usepackage[hyphens]{url}  
\usepackage{graphicx} 
\usepackage{natbib}  
\usepackage{caption} 
\usepackage{amsmath}
\usepackage{booktabs}
\usepackage{graphicx}
\usepackage{subcaption}
\usepackage{caption}
\usepackage{float}

\usepackage{algorithm}
\usepackage{algorithmic}

\usepackage{newfloat}
\usepackage{listings}
\DeclareCaptionStyle{ruled}{labelfont=normalfont,labelsep=colon,strut=off} 
\floatstyle{ruled}
\newfloat{listing}{tb}{lst}{}
\floatname{listing}{Listing}
\title{Know your Trajectory - Trustworthy Reinforcement Learning deployment through Importance-Based Trajectory Analysis}
\author{
    Clifford F\textsuperscript{1},
    Devika Jay\textsuperscript{1},
    Abhishek Sarkar\textsuperscript{2},
    Satheesh K Perepu\textsuperscript{2},\\
    Santhosh G S\textsuperscript{1},
    Kaushik Dey\textsuperscript{2},
    Balaraman Ravindran\textsuperscript{1}
}
\affiliations{
    \textsuperscript{1}Centre for Responsible AI, IIT Madras, India\\
    \textsuperscript{2}Ericsson Research, Bangalore, India

     na20b014@smail.iitm.ac.in, devika@cerai.in, abhishek.sarkar@ericsson.com, perepu.satheesh.kumar@ericsson.com,\\ santhoshgs013@gmail.com, deykaushik@ericsson.com, ravi@dsai.iitm.ac.in

}

\usepackage{bibentry}

\begin{document}

\maketitle

\begin{abstract}
As Reinforcement Learning (RL) agents are increasingly deployed in real-world applications, ensuring their behavior is transparent and trustworthy is paramount. A key component of trust is explainability, yet much of the work in Explainable RL (XRL) focuses on local, single-step decisions. This paper addresses the critical need for explaining an agent's long-term behavior through trajectory-level analysis. We introduce a novel framework that ranks entire trajectories by defining and aggregating a new state-importance metric. This metric combines the classic Q-value difference with a ``radical term'' that captures the agent's affinity to reach its goal, providing a more nuanced measure of state criticality. We demonstrate that our method successfully identifies optimal trajectories from a heterogeneous collection of agent experiences. Furthermore, by generating counterfactual rollouts from critical states within these trajectories, we show that the agent's chosen path is robustly superior to alternatives, thereby providing a powerful ``Why this, and not that?'' explanation. Our experiments in standard OpenAI Gym environments validate that our proposed importance metric is more effective at identifying optimal behaviors compared to classic approaches, offering a significant step towards trustworthy autonomous systems.\footnote{Code and set up to reproduce our experiments are available at: \url{https://github.com/clif-ford/XRL_Codebase}}\\
\end{abstract}


\section{Introduction}
\label{sec:introduction}

The increasing sophistication of Reinforcement Learning (RL) has enabled the training of agents for complex tasks, accelerating their deployment in diverse real-world systems. However, for these autonomous agents to be deemed trustworthy and responsible, their decision-making processes must be explainable. The field of Explainable RL (XRL) aims to provide high-fidelity, human-comprehensible explanations for an agent's behavior.

While a significant portion of XRL research has concentrated on local explanations, justifying a specific action in a given state \cite{10.1609/aaai.v38i9.28863}, these methods fall short of clarifying an agent's long-term strategy. Understanding the overarching ``story'' of an agent's behavior, encapsulated by its trajectory, is crucial for deployment in safety-critical domains. For instance, knowing why a self-driving car chose a particular route over another is more informative than knowing why it braked at a single intersection.

To address this gap, we propose a framework for explaining entire trajectories by leveraging causal inference concepts. Our approach produces explanations by answering contrastive questions like ``Why was this path taken?'' and ``Why was an alternative path not taken?''. Answering such questions involves generating counterfactuals, what could have happened, which is a powerful tool for human-like reasoning.

\begin{figure}[t]
 \includegraphics[width=0.5\linewidth]{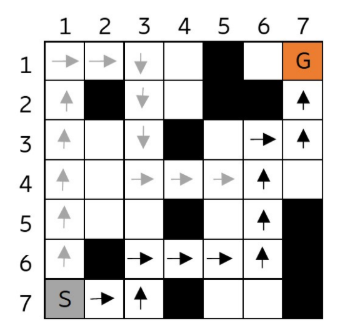}
    \centering
    \caption{An agent's observed trajectory (black) versus a longer, suboptimal alternative (gray). Our goal is to explain why the black path was chosen by demonstrating its superior importance.}
    \label{fig:grid}
\end{figure}

This paper makes the following contributions:
\begin{itemize}
    \item We introduce a novel state-importance metric that augments the standard Q-value difference with a ``radical term'' representing the agent's goal affinity, allowing for a more robust evaluation of state criticality.
    \item We propose a complete pipeline to rank entire trajectories by aggregating our state-importance metric, enabling the identification of the most salient and representative behaviors from a large dataset of experiences.
    \item We empirically validate our approach by generating counterfactuals from top-ranked trajectories, demonstrating that our method effectively highlights the optimality of the agent's chosen path comparing viable alternatives.
\end{itemize}

This work advances the state of deployable AI by providing a practical method for generating high-level strategic explanations, fostering greater trust and transparency in complex autonomous systems.

\section{Related Work}
\label{sec:litrev}

Explainable RL (XRL) is a well developing field, with methods broadly categorized into Feature Importance (FI), Learning Process and MDP (LPM), and Policy Level (PL) explanations \cite{10.1145/3616864}. Our work falls under the PL category, with a specific focus on trajectory-level explanations.

PL methods aim to explain high-level policy decisions. This includes summarizing key transitions \cite{10.5555/3237383.3237869}, converting complex recurrent policies into interpretable formats like finite state automata \cite{pmlr-v139-danesh21a}, or extracting prototypical ``landmark'' states from experience \cite{ef4d4b88353045fc812d88e9d1c1297c}.

Several works have specifically targeted trajectory explanations. The HIGHLIGHTS method \cite{10.5555/3237383.3237869} provides summaries by selecting states with the highest potential impact on future rewards, based on Q-values. While effective, it summarizes behavior through discrete states rather than analyzing the full trajectory sequence. Other approaches have used offline data to cluster trajectories and train surrogate policies to identify dissimilarities, attributing importance to clusters that cause the largest policy divergence \cite{deshmukh2024explainingrldecisionstrajectories}. However, the interpretability of these clusters can be a challenge. \cite{frost2022explainingreinforcementlearningpolicies} uses a learned policy to answer counterfactual queries by performing rollouts from matched states in a source domain, presenting these what-if scenarios to a user. More recently, visualization techniques have been used to abstract trajectories by clustering latent state representations, helping to visualize major state transitions for non-experts \cite{takagi2024abstractedtrajectoryvisualizationexplainability}.

Our work builds on these ideas by proposing a novel, principled metric for quantifying trajectory importance directly from the agent's value function. Unlike methods that rely on clustering or summarizing discrete states, our approach evaluates the entire sequential path, providing a holistic and contrastive explanation through counterfactual analysis.

\begin{figure*}[t]
\centering
\begin{subfigure}{0.4\textwidth}
    \centering
    \includegraphics[width=\linewidth]{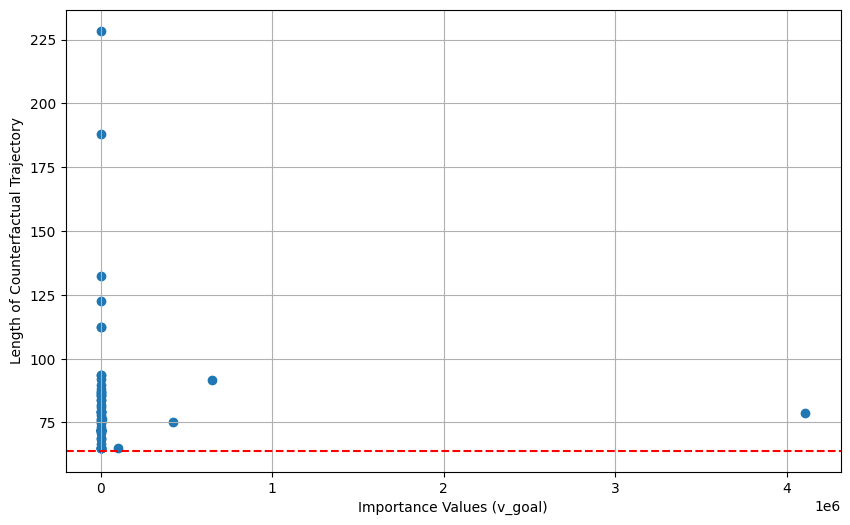}
    \caption{Our Method (V-Goal)}
    \label{fig:subfigA}
\end{subfigure}
\hfill
\begin{subfigure}{0.4\textwidth}
    \centering
    \includegraphics[width=\linewidth]{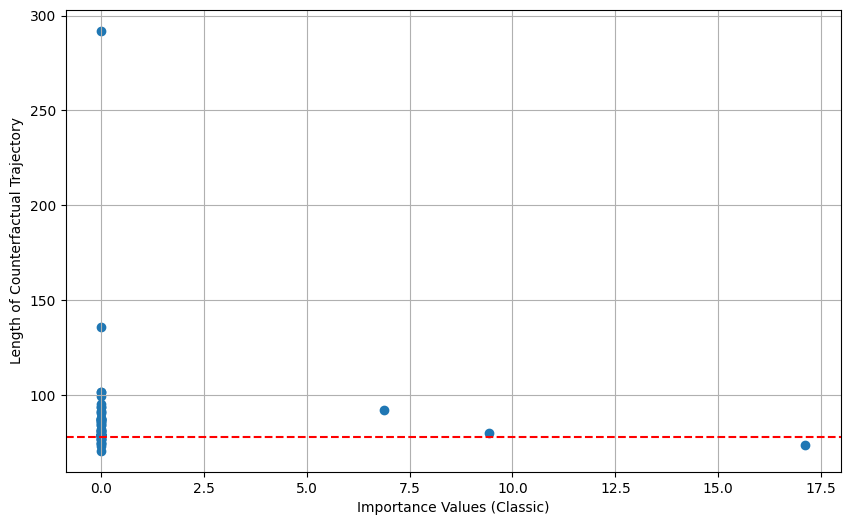}
    \caption{Classic Method ($\Delta Q$)}
    \label{fig:subfigB}
\end{subfigure}
\caption{Acrobot counterfactual trajectory lengths. The red line is the original trajectory's length. (a) For our method, all counterfactuals are longer (worse) than the original. (b) For the classic method, some counterfactuals are shorter (better), indicating it did not select a truly optimal trajectory to explain.}
\label{fig:results_acrobot_cf}
\end{figure*}

\section{Methodology}
\label{sec:method}

Our framework is designed to identify and explain the most significant trajectories from an agent's experience. The core of our approach lies in a novel definition of state and trajectory importance.

\subsection{State Importance: A Classic View}
We begin with the intuitive notion that a state is important if the choice of action within it has a significant impact on future rewards. This is classically defined using the agent's Q-values \cite{10.5555/3060621.3060733}. The importance of a state $s$, denoted $I(s)$, is the difference between the values of the best and worst possible actions:
\[
I(s) = \max_a Q^{\pi}(s, a) - \min_a Q^{\pi}(s, a)
\]
This quantity, $\Delta Q(s)$, captures the potential advantage available in state $s$. A high value indicates a critical decision point where a suboptimal action can be costly.

\subsection{A Modified Importance Metric}
While $\Delta Q(s)$ measures the potential gain, it does not capture the agent's confidence or decisiveness in pursuing the optimal action. A state might have a high $\Delta Q(s)$, but if the agent's policy is nearly uniform over several good actions, the state is less critical than one where a single action is decisively superior.

To address this, we introduce a modified state-action importance metric that incorporates the agent's affinity for reaching the goal. We define this as:
\[
I(s,a) = \Delta Q(s) \times R(s,a)
\]
Here, $R(s,a)$ is a ``radical term'' that quantifies the agent's commitment to its chosen path. We explored several formulations for $R(s,a)$:

\begin{enumerate}
    \item \textbf{Normalization (Naive)}: Measures how much better the chosen action is relative to the average action: $r(s,a) = (Q(s, a) - \mu_{Q}(s)) / \sigma_{Q}(s)$.
    \item \textbf{Bellman Error}: Uses the temporal difference error, $|Q(s, a) - (r + \gamma Q(s', a'))|$, as a measure of deviation from optimality.
    \item \textbf{Entropy-Based Confidence}: Measures the decisiveness of the policy $\pi(a|s)$. We define confidence as normalized negative entropy: $r(s) = 1 - (H(\pi(s)) / \log |\mathcal{A}|)$, where $r(s) \to 1$ for a deterministic policy.
    \item \textbf{Value-Based Goal Proximity}: Uses the state-value function $V(s)$ as a proxy for closeness to the goal. This can be normalized using a known range, $r(s) = (V(s) - V_{\min})/(V_{\max} - V_{\min})$, or relative to the goal state's value, $r(s) = |V(s)/V(s_{\text{final}})|$.
\end{enumerate}

Through experimentation, we found that the value-based goal proximity metric ($V_{\text{goal}}$) provided the most consistent and meaningful results, as it directly encodes progress towards the task objective.

\subsection{Trajectory Importance and Explanation}
To evaluate an entire trajectory, we aggregate the importance scores of its constituent state-action pairs. For a trajectory $\tau = \{(s_0, a_0), (s_1, a_1), \ldots, (s_T, a_T)\}$, its importance is the average state-action importance:
\[
I_\tau = \frac{1}{|\tau|} \sum_{(s, a) \in \tau} I(s,a) = \frac{1}{|\tau|} \sum_{(s, a) \in \tau} \Delta Q(s) \times R(s,a)
\]
This score allows us to rank a large collection of trajectories, identifying those that are most representative of the agent's optimal strategy.

\subsection{Explanation Pipeline}
Our full pipeline for generating trajectory explanations is as follows:
\begin{enumerate}
    \item \textbf{Data Collection}: Collect a dataset of trajectories and populate a Q-table from a trained agent's critic. For continuous state spaces, we discretize the state representations.
    \item \textbf{Importance Calculation}: For each state-action pair $(s, a)$ in every trajectory, calculate our modified importance metric $I(s,a)$.
    \item \textbf{Trajectory Ranking}: Compute the aggregate importance $I_\tau$ for each trajectory and rank them to find the top-$k$ most important ones. We select the trajectory from this set with the best outcome (e.g., highest reward, shortest length).
    \item \textbf{Counterfactual Generation}: For the top-ranked trajectory, generate counterfactuals. At each state $s_i$ along the original path, we forbid the original action $a_i$ and force the agent to take a different action, after which it follows its policy $\pi$. This produces a set of alternative trajectories.
    \item \textbf{Contrastive Explanation}: Compare the original trajectory with the generated counterfactuals on metrics like total reward, length, and importance score. An optimal original trajectory should be demonstrably better than its counterfactuals, providing a powerful explanation for the agent's behavior.
\end{enumerate}

This pipeline provides a concrete method for answering ``Why this path and not another?'' by showing the consequences of deviation.

\begin{figure*}[t]
\centering
\begin{subfigure}{0.4\textwidth}
    \centering
    \includegraphics[width=\linewidth]{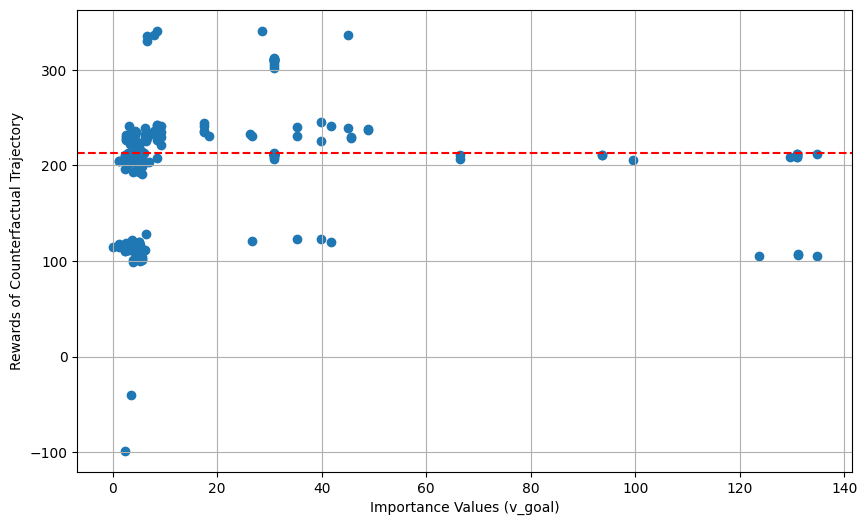}
    \caption{Our Method (V-Goal)}
    \label{fig:subfigE}
\end{subfigure}
\hfill
\begin{subfigure}{0.4\textwidth}
    \centering
    \includegraphics[width=\linewidth]{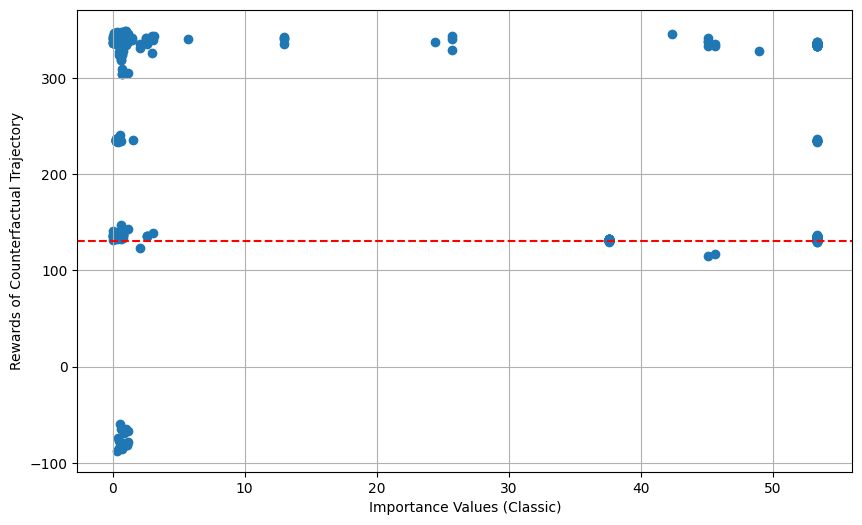}
    \caption{Classic Method ($\Delta Q$)}
    \label{fig:subfigF}
\end{subfigure}
\caption{LunarLander counterfactual trajectory rewards. The red line represents the original trajectory's reward. (a) For our method, all counterfactuals yield lower rewards. (b) For the classic method, some counterfactuals result in higher rewards.}
\label{fig:results_lunar_cf}
\end{figure*}

\section{Experiments and Results}
\label{sec:res}

We conducted experiments in OpenAI Gym environments \cite{brockman2016openaigym}, Acrobot-v1 and LunarLander-v2, using agents trained with the PPO algorithm. We focused on the scenario where trajectories are collected throughout the training process, resulting in a dataset containing both optimal and suboptimal behaviors. Our framework must be able to distinguish between them.

\subsection{Identifying Optimal Trajectories}
We first evaluated the ability of different radical terms ($R(s,a)$) to identify the best trajectories. For each metric, we ranked all collected trajectories and computed the average length and reward of the top 5.

\begin{table}[h]
\centering
\scalebox{0.9}{
\begin{tabular}{l|cc}
\toprule
\textbf{Method} & \textbf{Avg. Length} & \textbf{Avg. Reward}\\
\midrule
\multicolumn{3}{c}{\textit{Acrobot-v1 Environment}} \\
\hline
Classic ($\Delta Q$ only) & 70.0 & -69.0 \\
Naive Normalization & 70.0 & -69.0 \\
Entropy-Based & 73.2 & -72.2 \\
Bellman Error & 70.8 & -69.8 \\
V-Normalization & 70.0 & -69.0 \\
\textbf{V-Goal} & \textbf{68.8} & \textbf{-67.8} \\
\bottomrule
\end{tabular}}
\caption{Performance of top-5 ranked trajectories in Acrobot. Lower length and higher (less negative) reward are better. Our \textbf{V-Goal} metric identifies the most optimal set of trajectories.}
\label{tab:result_acrobot}
\end{table}

The results for the Acrobot environment are shown in Table \ref{tab:result_acrobot}. In this task, success is measured by achieving the goal in the fewest steps, resulting in a higher (less negative) reward. While the differences are subtle, the trajectories ranked highest by our `V-Goal' metric are consistently the most efficient, with the shortest average length (68.8) and highest average reward (-67.8). This provides initial evidence that incorporating goal affinity helps refine the selection of optimal trajectories. The distinction becomes much clearer in the more complex LunarLander environment.

\begin{table}[h]
\centering
\scalebox{0.9}{
\begin{tabular}{l|cc}
\toprule
\textbf{Method} & \textbf{Avg. Reward} & \textbf{Avg. Length}\\
\midrule
\multicolumn{3}{c}{\textit{LunarLander-v2 Environment}} \\
\hline
Classic ($\Delta Q$ only) & 116.87 & 1000.0 \\
Bellman Error & 117.37 & 1000.0 \\
Naive Normalization & 188.12 & 433.2 \\
Entropy-Based & 121.27 & 871.0 \\
V-Normalization & 120.59 & 1000.0 \\
\textbf{V-Goal} & \textbf{207.13} & \textbf{319.2} \\
\bottomrule
\end{tabular}}
\caption{Performance of top-5 ranked trajectories in LunarLander. Higher reward and lower length are better. The results starkly highlight the effectiveness of our proposed metric.}
\label{tab:result_lunar}
\end{table}

The data in Table \ref{tab:result_lunar} clearly shows the superiority of the V-Goal' metric. In LunarLander, a successful landing yields a high reward, while crashing or running out of time (max 1000 steps) results in poor scores. Our `V-Goal' metric is the only method that consistently identifies successful landing trajectories, achieving an average reward over 200 and an average trajectory length of 319 steps. In contrast, the classic method and others select trajectories that hit the time limit, indicating failed or meandering attempts. This suggests that incorporating goal proximity via the value function is essential for distinguishing truly optimal, task-achieving behavior from prolonged, suboptimal attempts.

\subsection{Counterfactual Explanations}

Having established `V-Goal' as our best metric, we generated counterfactuals for the single best trajectory it identified and compared them with those from the top trajectory identified by the classic $\Delta Q$ metric. A successful explanation would show that deviations from the original trajectory lead to worse outcomes (e.g., longer paths, lower rewards).

Figure \ref{fig:results_acrobot_cf} shows the results for Acrobot. For the trajectory selected by our `V-Goal' metric (Fig. \ref{fig:subfigA}), every generated counterfactual trajectory was longer (worse) than the original. This provides a strong, clear explanation: the agent followed the optimal path, and any deviation would have been suboptimal. In contrast, for the trajectory selected by the classic $\Delta Q$ metric (Fig. \ref{fig:subfigB}), several counterfactuals were shorter than the original, indicating that the classic method failed to identify a truly optimal trajectory, thus providing a confusing or incorrect explanation.

We observed the same pattern for LunarLander, shown in Figure \ref{fig:results_lunar_cf}. Counterfactuals from the trajectory identified by our method consistently resulted in lower rewards, while the classic method again selected a trajectory from which better paths could be found. These results robustly demonstrate that our modified importance metric is superior for identifying and explaining optimal agent behavior.

\subsection{Discussion}
Our experiments demonstrate that when an agent is still learning, with an experience buffer containing both successful and suboptimal trajectories, our framework excels at identifying and explaining optimal behavior.

However, when analyzing trajectories generated exclusively by a fully trained, optimal agent, the task becomes more challenging. In this scenario, most trajectories are nearly identical in quality, offering less explanatory insight. Future work could address this by focusing not on ranking trajectories but on identifying the few most \textbf{critical states} within a single optimal trajectory. Generating counterfactuals from only these pivotal moments could provide more concise and impactful explanations, even for highly optimized agents, maintaining our focus on trajectory-level analysis while adapting to the nuances of expert behavior.

\section{Conclusion and Future Work}
\label{sec:final}

In this paper, we introduced a framework for generating trajectory-level explanations in Reinforcement Learning. By defining a state-importance metric that accounts for both Q-value advantage and goal affinity, our method identifies and ranks optimal trajectories from heterogeneous experience data. Through contrastive counterfactuals, we provide clear, intuitive explanations for an agent's long-term strategy, demonstrating why its chosen path was superior to alternatives.

Our empirical results show that this approach is more effective than classic importance metrics, providing a more reliable foundation for trustworthy and deployable AI systems. Understanding and interrogating high-level behavior is a critical step towards deploying RL safely in the real world.

For future work, we plan to extend this framework to automatically identify critical decision points within trajectories and explore scenarios where the agent's policy and value functions are unknown. Techniques from Inverse Reinforcement Learning (IRL) could infer a reward function explaining observed trajectories, after which our importance-based analysis can explain the inferred policy.

\bibliography{aaai2026}

@article{10.1145/3616864,
author = {Milani, Stephanie and Topin, Nicholay and Veloso, Manuela and Fang, Fei},
title = {Explainable Reinforcement Learning: A Survey and Comparative Review},
year = {2024},
issue_date = {July 2024},
publisher = {Association for Computing Machinery},
address = {New York, NY, USA},
volume = {56},
number = {7},
issn = {0360-0300},
url = {https://doi.org/10.1145/3616864},
doi = {10.1145/3616864},
journal = {ACM Comput. Surv.},
month = apr,
articleno = {168},
numpages = {36}
}

@inproceedings{10.5555/3237383.3237869,
author = {Amir, Dan and Amir, Ofra},
title = {HIGHLIGHTS: Summarizing Agent Behavior to People},
year = {2018},
publisher = {International Foundation for Autonomous Agents and Multiagent Systems},
address = {Richland, SC},
booktitle = {Proceedings of the 17th International Conference on Autonomous Agents and MultiAgent Systems},
pages = {1168–1176},
numpages = {9},
location = {Stockholm, Sweden},
series = {AAMAS '18}
}

@misc{deshmukh2024explainingrldecisionstrajectories,
      title={Explaining RL Decisions with Trajectories}, 
      author={Shripad Vilasrao Deshmukh and Arpan Dasgupta and Balaji Krishnamurthy and Nan Jiang and Chirag Agarwal and Georgios Theocharous and Jayakumar Subramanian},
      year={2024},
      eprint={2305.04073},
      archivePrefix={arXiv},
      primaryClass={cs.AI},
      url={https://arxiv.org/abs/2305.04073}, 
}

@misc{takagi2024abstractedtrajectoryvisualizationexplainability,
      title={Abstracted Trajectory Visualization for Explainability in Reinforcement Learning}, 
      author={Yoshiki Takagi and Roderick Tabalba and Nurit Kirshenbaum and Jason Leigh},
      year={2024},
      eprint={2402.07928},
      archivePrefix={arXiv},
      primaryClass={cs.HC},
      url={https://arxiv.org/abs/2402.07928}, 
}

@misc{frost2022explainingreinforcementlearningpolicies,
      title={Explaining Reinforcement Learning Policies through Counterfactual Trajectories}, 
      author={Julius Frost and Olivia Watkins and Eric Weiner and Pieter Abbeel and Trevor Darrell and Bryan Plummer and Kate Saenko},
      year={2022},
      eprint={2201.12462},
      archivePrefix={arXiv},
      primaryClass={cs.LG},
      url={https://arxiv.org/abs/2201.12462}, 
}

@inproceedings{10.5555/3060621.3060733,
author = {Amir, Ofra and Kamar, Ece and Kolobov, Andrey and Grosz, Barbara J.},
title = {Interactive teaching strategies for agent training},
year = {2016},
isbn = {9781577357704},
publisher = {AAAI Press},
booktitle = {Proceedings of the Twenty-Fifth International Joint Conference on Artificial Intelligence},
pages = {804–811},
numpages = {8},
location = {New York, New York, USA},
series = {IJCAI'16}
}

@inproceedings{10.1609/aaai.v38i9.28863,
author = {Amitai, Yotam and Septon, Yael and Amir, Ofra},
title = {Explaining reinforcement learning agents through counterfactual action outcomes},
year = {2024},
isbn = {978-1-57735-887-9},
publisher = {AAAI Press},
url = {https://doi.org/10.1609/aaai.v38i9.28863},
doi = {10.1609/aaai.v38i9.28863},
booktitle = {Proceedings of the Thirty-Eighth AAAI Conference on Artificial Intelligence and Thirty-Sixth Conference on Innovative Applications of Artificial Intelligence and Fourteenth Symposium on Educational Advances in Artificial Intelligence},
articleno = {1115},
numpages = {9},
series = {AAAI'24/IAAI'24/EAAI'24}
}

@InProceedings{pmlr-v139-danesh21a,
  title = 	 {Re-understanding Finite-State Representations of Recurrent Policy Networks},
  author =       {Danesh, Mohamad H and Koul, Anurag and Fern, Alan and Khorram, Saeed},
  booktitle = 	 {Proceedings of the 38th International Conference on Machine Learning},
  pages = 	 {2388--2397},
  year = 	 {2021},
  editor = 	 {Meila, Marina and Zhang, Tong},
  volume = 	 {139},
  series = 	 {Proceedings of Machine Learning Research},
  month = 	 {18--24 Jul},
  publisher =    {PMLR},
  url = 	 {https://proceedings.mlr.press/v139/danesh21a.html}
}

@inproceedings{ef4d4b88353045fc812d88e9d1c1297c,
title = "CAPS: Comprehensible Abstract Policy Summaries for Explaining Reinforcement Learning Agents",
author = "Joe McCalmon and Thai Le and Sarra Alqahtani and Dongwon Lee",
note = "Publisher Copyright: {\textcopyright} 2022 International Foundation for Autonomous Agents and Multiagent Systems (www.ifaamas.org). All rights reserved; 21st International Conference on Autonomous Agents and Multiagent Systems, AAMAS 2022 ; Conference date: 09-05-2022 Through 13-05-2022",
year = "2022",
language = "English (US)",
series = "Proceedings of the International Joint Conference on Autonomous Agents and Multiagent Systems, AAMAS",
publisher = "International Foundation for Autonomous Agents and Multiagent Systems (IFAAMAS)",
pages = "889--897",
booktitle = "International Conference on Autonomous Agents and Multiagent Systems, AAMAS 2022",
}

@misc{brockman2016openaigym,
      title={OpenAI Gym}, 
      author={Greg Brockman and Vicki Cheung and Ludwig Pettersson and Jonas Schneider and John Schulman and Jie Tang and Wojciech Zaremba},
      year={2016},
      eprint={1606.01540},
      archivePrefix={arXiv},
      primaryClass={cs.LG},
      url={https://arxiv.org/abs/1606.01540}, 
}

\appendix
\section{Appendix}

\subsection{Supplementary Counterfactual Results}
\begin{figure*}[t]
\centering
\begin{subfigure}{0.43\textwidth}
    \centering
    \includegraphics[width=\linewidth]{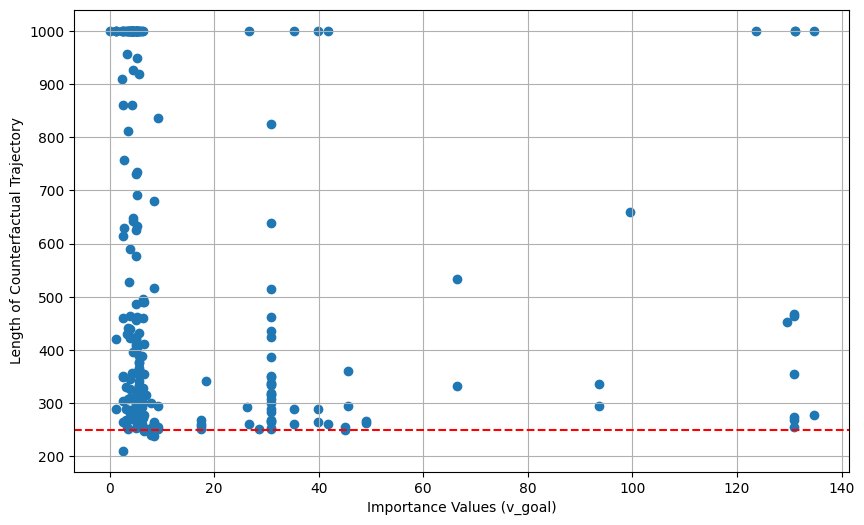}
    \caption{Our Method (V-Goal)}
    \label{fig:subfigC}
\end{subfigure}
\hfill
\begin{subfigure}{0.43\textwidth}
    \centering
    \includegraphics[width=\linewidth]{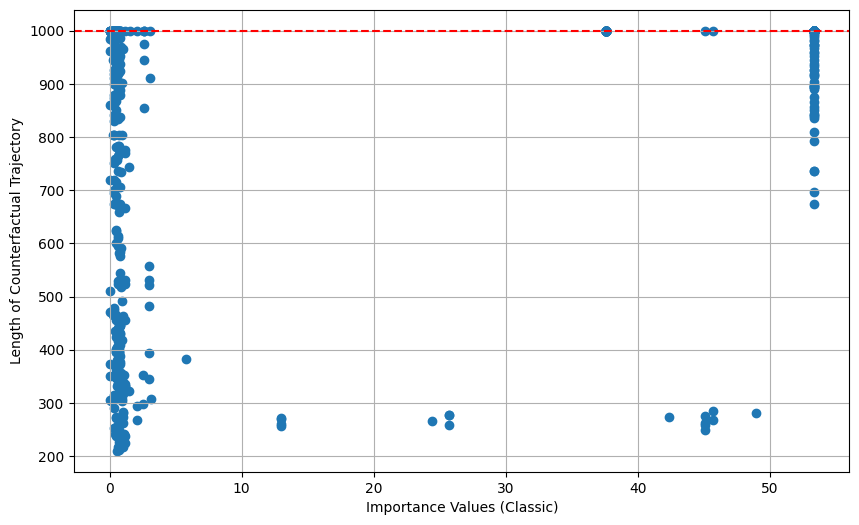}
    \caption{Classic Method ($\Delta Q$)}
    \label{fig:subfigD}
\end{subfigure}
\caption{LunarLander counterfactual trajectory lengths. The red line represents the original trajectory's length. (a) Counterfactuals from our method’s selected trajectory are probabilistically longer. (b) The classic method’s selected trajectory has counterfactuals that are probabilistically shorter.}
\label{fig:results_lunar_cf_length}
\end{figure*}

As discussed in the main paper's counterfactual analysis, our `V-Goal' metric consistently identifies trajectories that are superior to their alternatives. The main text demonstrated this using reward values and trajectory length for LunarLander (Figure \ref{tab:result_lunar}). This appendix provides supplementary evidence using trajectory length as the metric (Figure \ref{fig:results_lunar_cf_length}), further reinforcing our claim that the identified optimal trajectory is robustly better than its counterfactuals across multiple performance dimensions.

\subsection{Exploration of KL-Divergence Metric}
During our research, we explored using KL-divergence as a radical term, $r(s) = KL(\pi(s), X(s))$, where $\pi(s)$ is the agent's policy and $X(s)$ is a reference distribution over actions. The intuition was that a high divergence would indicate a confident, non-uniform policy. However, this metric was ultimately not included in our final framework for two main reasons:
\begin{enumerate}
    \item \textbf{High Variance}: The performance was highly sensitive to the choice of the reference distribution $X(s)$ and varied significantly across different environments.
    \item \textbf{Lack of Clear Rationale}: It was difficult to establish a principled, general method for choosing the reference distribution $X(s)$ for any given environment. While we hypothesized certain distributions might be suitable for specific agent behaviors (as shown in Table \ref{tab:KL}), this could not be consistently validated.
\end{enumerate}

Due to this lack of stability and clear justification, we concluded that the KL-divergence metric was not robust enough for a general-purpose explanation framework.

\begin{table*}[h]
\centering
\renewcommand{\arraystretch}{1.3} 
\setlength{\tabcolsep}{6pt}      
\begin{tabular}{|c|p{0.75\linewidth}|}
\hline
\textbf{Distribution} & \textbf{When to Use} \\ \hline \hline
\textbf{Uniform} &
When no clear preference exists in action selection; suitable for exploratory, early-stage agents or when the action tendencies are uncertain. \\ \hline

\textbf{Gaussian} &
When actions follow a central tendency with some variability (common in continuous action spaces). Ideal for confident, near-deterministic agents. \\ \hline

\textbf{Exponential} &
When large actions are rare and small actions are frequent (e.g., sparse high-reward events). Suitable for exploitative agents. \\ \hline

\textbf{Dirichlet} &
When some actions are preferred over others but there remains significant variability. Useful for environments with multiple viable paths to success. \\ \hline

\textbf{Beta} &
When actions have bounded probabilities (0–1) and model uncertainty in preference; suitable for tasks balancing exploration and exploitation. \\ \hline
\end{tabular}
\caption{Initial hypotheses for choosing a reference distribution $X(s)$ for the KL-divergence metric.}
\label{tab:KL}
\end{table*}


\end{document}